# Data Center Audio/Video Intelligence on Device (DAVID) - An Edge-AI Platform for Smart-Toys

Gabriel Cosache
*DTS (office of CTO),*
*Xperi Corporation*
Galway, Ireland
gabriel.costache@xperi.com

Francisco Salgado
*DTS (office of CTO),*
*Xperi Corporation*
Galway, Ireland
line 5: email address or ORCID

Rishabh Jain
*C3 Imaging Research Center*
*University of Galway*
Galway, Ireland
rishabh.jain@universityofgalway.ie

Cosmin Rotariu
*DTS (office of CTO),*
*Xperi Corporation*
Galway, Ireland
cosmin.rotariu@xperi.com

George Sterpu
*DTS (office of CTO),*
*Xperi Corporation*
Galway, Ireland
george.sterpu@xperi.com

Peter Corcoran
*C3 Imaging Research Center*
*University of Galway*
Galway, Ireland
peter.corcoran@universityofgalway.ie

*Abstract*—An overview is given of the DAVID Smart-Toy platform, one of the first Edge-AI platform designs to *incorporate* advanced low-power data processing by neural inference models co-located with the relevant image or audio sensors. There is also on-board capability for in-device text-to-speech generation. Two alternative embodiments are presented – a smart Teddy-bear and a roving dog-like robot. The platform offers a speech-driven user interface and can observe and interpret user actions and facial expressions via its computer vision sensor node. A particular benefit of this design is that no personally identifiable information passes beyond the neural inference nodes thus providing inbuild compliance with data protection regulations.

## I. Introduction

Edge-AI is an emerging concept implying a migration of computational intelligence and associated data processing from cloud repositories to occur closer to the source of data. In some interpretations it implies data processing on a local smartphone or hub device, but the ultimate goal of Edge-AI is on-device processing. Typically, the artificial intelligence element is a neural model that leverages recent advances in processing image, speech, or other raw sensor data sources. However, as the capabilities of most embedded inference chipsets are relatively limited and still require significant compute power [1], [2] most designs implement quite limited or specific functionality [3]. The inference requirements of advanced computer vision and automated speech models further limit the capabilities of Edge-AI implementations [2], [4]–[6]. Fortunately a new generation of specialized neural accelerators [7]–[9] are capable of running larger neural models and even combinations of models, as we shall see.

Data privacy has also become a significant consideration, especially for consumer devices and services [10]–[13]. More specifically, smart-toys have led to much controversy when they collect personally identifiable data from children [14], [15]. Clearly, privacy is of particular importance when dealing with children. Thus, data security has been a primary concern for the platform and a key design requirement was to eliminate sending any data that might be considered as personally identifiable beyond the sensor nodes. Due to the scope of the *General Data Protection Regulations* (GDPR) in Europe [16] this implies that all image or speech data should be processed on the sensor boards.

## II. System Design and Architecture Overview

The DAVID project required a large-scale development effort over a three-year period, so it is not feasible to capture the many details of the design, test and final implementation of the hardware designs. Here we present the final working system, focusing on the key privacy-by-design aspects.

### A. The ERGO Neural Accelerator

The DAVID platform design was inspired by the recent availability of low-power inference chipset such as ERGO [17]. It provides ultra-low-power capabilities while delivering significant computational capabilities – of the order of 50 TOPS/Watt. Thus, for the smart-toy use case a computational loading of 2-3 TOPS is feasible for a power budget of 50 mW – approximately the same power consumption as two light-emitting diodes.

### B. Systen Architecture Overview

The system hardware layout is provided in *Figure 1*. This comprises three dedicated inference nodes, one with an onboard camera enabling a computer vision node; the second with a microphone to provide an audio sensing node – the primary function will be operating the speech interface with the user, and the third board is linked to a speaker to enable the smart toy to generate a neural voice output. This third board is not a sensor board, but is needed to close the loop on a speech-based user interface. This demonstrates the capability of Edge-AI to also generate data outputs.

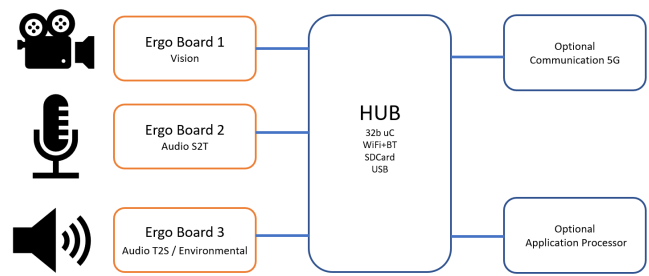

*Figure 1:* System Hardware Architecture for DAVID

These three inference boards are connected via an I2S bus to a central, low-power microcontroller (MCU) hub. This is also an ultra-low-power design, but consumes more power than the sensor nodes, unless placed in deep shutdown mode.

*C. The Hub Board*

The Hub board, shown in *Figure 2*, was designed to leverage the many state-of-the-art features available on today's smartphones including IMU subsystem, programmable wireless connectivity and JTAG. It features a STM32H7 low power MCU which can support full 32-bit OS. This was selected for his flexibility and support for all peripherals we have considered combined with its low power requirements. The Hub board's main purpose is to connect the Ergo boards and managing the communication. The Hub should only be live when the Ergo boards are sensing events that requires the Hub board to provide input to any of the peripheral attached to it. It will be in sleep mode most of the time. This helps to ensure very low power requirement for the overall system platform.

TABLE 1: CONNECTIVITY, MEMORY & COMPUTATIONAL CAPABILITIES OF THE DAVID PLATFORM (HUB + NODES)

| CONNECTIVITY | HUB | NODE | COMPUTATION | HUB | NODE |
|---|---|---|---|---|---|
| I2S (Tx, Rx), I2C | X | X | Ergo, 55TOPs/Watt x3 | | X |
| MIPI and Parallel | | X | Arc CPU\DSP) | | X |
| SPI & QSPI | X | X | STM32 (Arm M7) | X | |
| GPIO (32 bit) | X | X | ESP32 (Xtensa LX6) | X | |
| FTDI (JTAG, UART) | X | | | | |
| WiFi/BT | X | | | | |
| USB OTG | X | | | | |
| **Memory** | | | | | |
| 16MB QSPI Flash | | X | | | |
| 128MB QSPI Flash | X | | | | |
| 32MB SRAM | X | | | | |
| 448 KB ROM | ESP32 | | | | |
| 520 KB SRAM | ESP32 | | | | |
| SDCard | X | | | | |

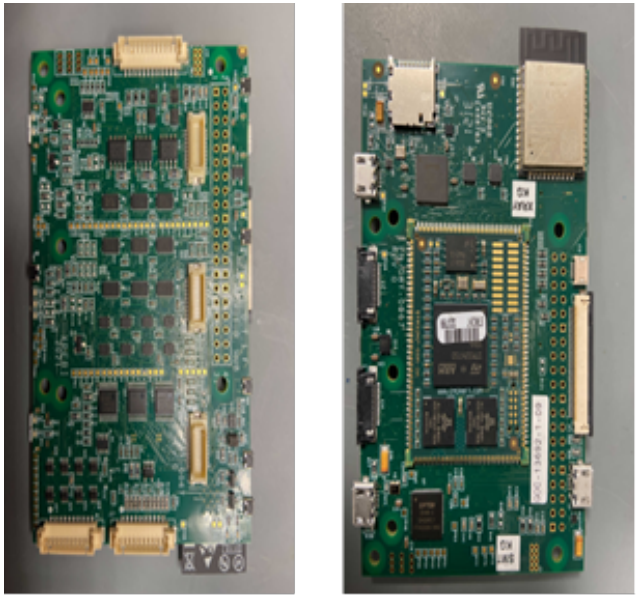

*Figure 2: Top and Bottom views of the DAVID Hub board.*

*D. The Sensor Nodes*

The sensor nodes are designed to leverage the capabilities of the ERGO chipset. They feature on-board MIPI bus to allow for fast real-time data transfer from an on-board camera or other high-bandwidth sensors and to allow fast data transfers between ERGO and the on-board memory subsystem. Each node also has a built-in IMU and I/O ports to support a range of different sensors peripherals. For the initial DAVID proof-of-concept (PoC) three different sensor nodes were configured:

**Vision Node:** This node features a wide-field QVGA MIPI camera. This choice was made as a trade-off between the resolution required to implement most of the computer vision algorithms and the complexity of the neural architectures. Most of the selected neural algorithms operate well with QVGA input, and this allows more neural models to be incorporated and operate in parallel. The MIPI architecture is scalable to a higher (or lower) resolution camera sensor as required.

**Audio Node:** This node features two microphone inputs – one is a low-quality consumer grade microphone, but a higher quality stereo microphone is also incorporated and used for initial demos and to gather test input data with different quality.

**Speaker Node:** This node was configured with a number of different digital output speakers for testing. From our experience it is difficult to achieve significant loudness in the output without added amplification, but this has consequences for power consumption. To date we did not find a good low power solution and the PoC speech output can be difficult to hear in a noisy environment. However, ongoing improvements in sound output components from smartphones should help solve this issue at a sensible price/performance point in the near future.

*E. The DAVID Platform*

The hub board and up to three Inference nodes are designed to be assembled into a single platform unit that can be designed into a proof-of-concept smart-toy. This is illustrated as block-diagram in *Figure 3, below*. Various on-board connectors simplify connecting the electronic platform to externally mounted cameras, microphones, sensors, speakers, or other equivalent peripherals that consume or generate data. The design is intended to be as generic as possible to facilitate incorporation into demonstrators for different consumer devices/products. A picture of the two sides of the hub board is shown in *Figure 2,* opposite..

Naturally this platform is not intended for mass-market manufacturing, only for proof-of-concept (PoC) designs. Ultimately the different system components would be incorporated into a single chip for a particular product design. Here the selection of ARM based MCU and mass-market camera, microphone and speaker peripherals will simplify the transition from PoC to final mass-market product. The fully assembled system is shown in *Figure 4,* below.

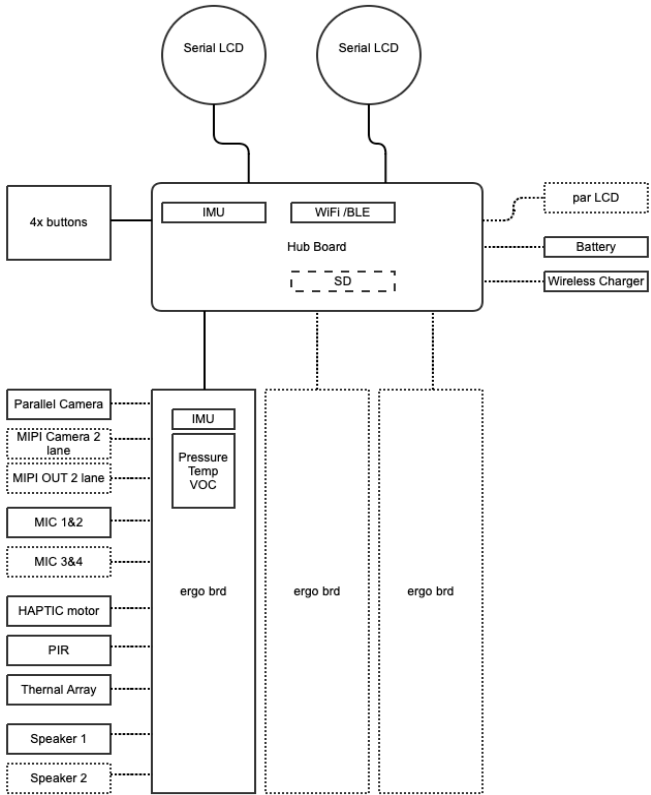

*Figure 3: Detailed System Block Diagram.*

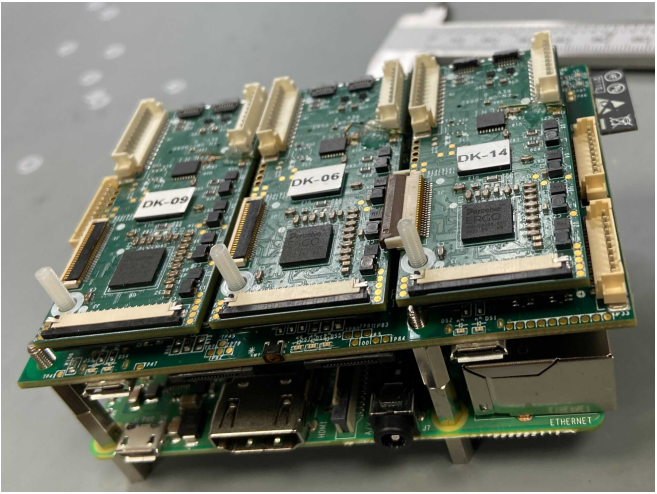

**Figure 4:** *DAVID Platform – Hub with 3 Inference node daughter boards fully assembled for placement in a smart-toy.*

### III. EDGE-AI MODELS AND THE SYSTEM USER INTERFACE

In this section we take a look at the neural models implemented on each of the node boards and discuss some of the challenges in training and preparing neural models to run on the DAVID platform.

*A. Porting of NN Models to the Edge-AI Platform*

Prior to compilation and compression of a neural model onto the ERGO is it important to simplify the original neural model (or composite model if several models are to combine on a single ERGO board). This is a complex process and typically involves a mix of layer quantization and node pruning. In some cases – a well-known example is the YOLO object detection framework – there may be a range of model sizes (tiny, small, regular, large models) with varying levels of performance/complexity to provide a repeatable starting point. Here, due to space limitations, we only comment that this process is empirical and there is a need to balance the desired levels of performance with the capabilities of the ERGO tools to covert models for loading on ERGO. In theory any model that can run on *PyTorch* can be converted, but in some cases, it was found that models simply could not achieve acceptable levels of performance.

*B. The Vision Node*

This sensing node is perhaps the most advanced as it leverages several decades of computational imaging experience in the CTO office of Xperi [18], [19]. This node incorporates a multi-function neural model that can perform a range of vision tasks. An overview of one representative version is shown in *Figure 5*, but this inference module can be configured in various alternative arrangements, depending on the specific application. Here the neural model can detect both facial, hand and body regions and a set of analytics is performed on each of these. In addition, body landmarks, hand gestures (from a fixed set of classes) and facial characteristics are output. The facial analysis is the most sophisticated, including orientation, a landmark mesh, facial expression (from a fixed set of classes) and facial embedding data that can be used to authenticate the user. All of these data outputs are available in numeric form to the hub board but no sensitive biometric data is exported beyond the vision node.

In addition, this example also includes a neural video encoder that allows an encoded video stream to be set over a secure wireless link (Bluetooth) to a parent's phone. The video encoding can only be viewed on a paired app running on the phone and requires a custom decoder to view the video. This is the only export of PID data outside of the platform and is provided to show that secure parental access can be provided. The functionality illustrated in *Figure 5* is available in real-time at frame rates of 30 fps. The total power cost is 100 mW. By deactivating the secure streaming functionality this can be reduced significantly. It is worth noting that the system architecture allows for fast reprogramming of each Inference node. Typically each can be re-flashed with a separate functionality in a few 10's of milliseconds. Thus this node could be re-flashed with a different set of functionality in order to support a specific play activity or to switch between single-player and multi-player use case. Multiple Inference models can be stored in flash, or uploaded via a secure app to the system allowing for additional flexibility in uses of the platform.

*C. The Text-to-Speech (TTS) Node*

The TTS Edge AI pipeline is composed of two modules: a spectrogram network and a vocoder. The spectrogram network was initially adapted from the well-known Tacotron model [20]. However, this architecture required off-node computations in the hub MCU and the model was slow to re-converge. Due to these challenges later versions of the TTS model switched to explore the FastSpeech end-to-end model [21], [22] and later several optimized versions of this model [23]. For the Vocoder, several alternatives were explored with the chosen architecture being HiFiGAN [24]. An overview of TTS architectures can be found here [20].

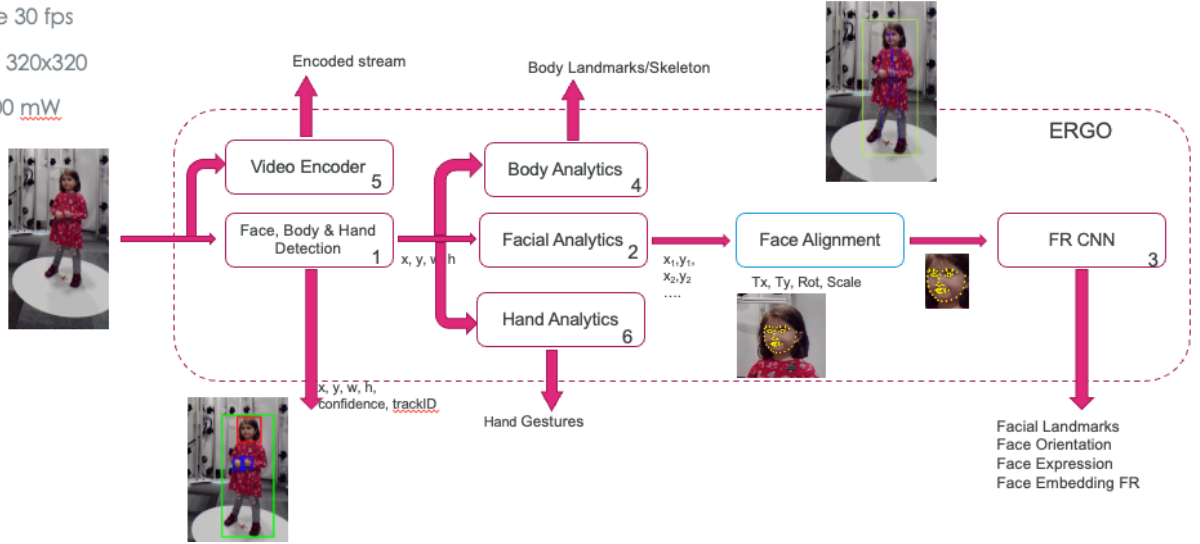

*Figure 5: The multi-CNN model structure implemented in the Computer Vision Sensor Node of DAVID.*

The optimized spectrogram network, shown in *Figure 6*, is a fully convolutional non-autoregressive network. It computes an intermediate log mel-scale spectrogram representation of the whole input text in parallel. The network takes the text characters as inputs, avoiding a separate phoneme conversion module. The network models the durations of the encoded characters directly prior to upsampling them to match the spectrogram length. During training, it uses the self-alignment module from FastPitch [21], which removes the need for external precomputed alignments between the text and the corresponding audio and easily scales to new datasets.

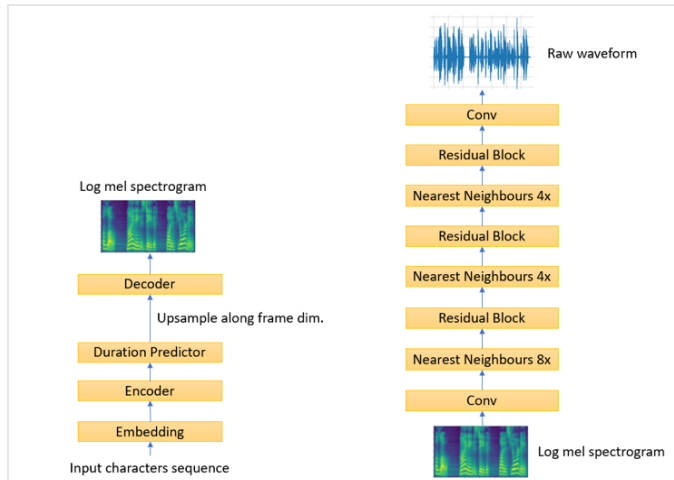

*Figure 6: Left-side: The Spectrogram network diagram. Right-side: The Vocoder generator diagram.*

The vocoder follows the generator architecture of [24] interleaving fully convolutional residual blocks with nearest neighbours upsampling layers. The convolutional kernel sizes and the receptive field of the network have been tuned for inference on the hardware at the chosen output sampling rate. During inference, the vocoder uses a sliding window approach for reducing activation memory. The network takes as input non overlapping parts of the log mel-scale spectrogram and computes the corresponding audio waveforms, which are then concatenated to produce the complete speech waveform.

Both modules are trained separately. The spectrogram network is trained by optimizing the mean-squared error between the log mel-scale spectrogram of the predicted and ground-truth signals as well as the alignment error. The vocoder is trained using loss proposed in [24].

### D. The Automatic Speech Recognition Node

SpeechNet is a fully convolutional neural architecture designed for real-time speech recognition on the ERGO hardware. In contrast with related convolutional models in ASR, such as Jasper [22] or QuartzNet [23] SpeechNet relies on a considerably shorter audio context, and uses small kernel sizes. The shorter receptive field is achieved by reducing the network depth at the expense of the layer width, where multiple convolutions are executed in parallel, similar to the building block of the Inception architecture [24].

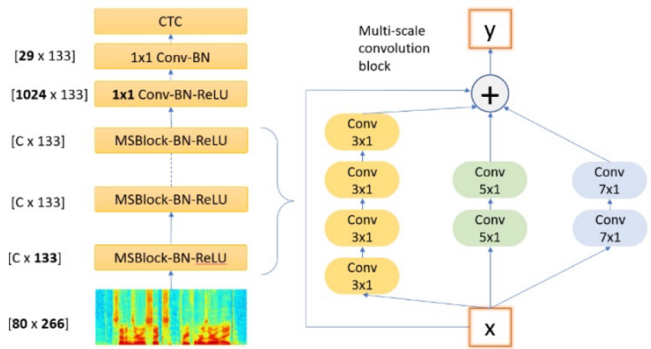

*Figure 7: SpeechNet1 diagram. Left: Overall structure. Right: Zoom on a Multi-scale convolution block (MSBlock)*

The basic building block of SpeechNet is shown on the right-hand side of the diagram in *Figure 7*, and consists of several Convolutional layers, each followed by Batch Normalization (BN) and Rectified Linear Unit (ReLU) non-linearities. For brevity, the BN and ReLU operations are not included in the diagram. The block input *x* is transformed by three different convolutional paths made of two or more

Conv-BN-ReLU blocks, each path using a different kernel size. Furthermore, with the exception of the first MSBlock in the network, all MSBlocks add a residual connection from input to output. The input feature to SpeechNet is a normalised time-frequency log mel-scale spectrogram representation of an audio signal. The model outputs an array of 29 posterior probabilities defined over an inventory of 26 letters in the English alphabet, blank space, apostrophe, and the CTC skip/BLANK token, and is trained using the Connectionist Temporal Classification (CTC) objective function [25] within a supervised learning framework.

For simplicity, SpeechNet uses characters as modelling units, lessening the burden of generating richer transcriptions. Through an appropriate parametrization of the network in relation to depth, kernel sizes, strides, and padding, SpeechNet1 achieves a constant algorithmic latency of 1.3 seconds, relying on 3.325 seconds of audio context. Building upon SpeechNet, we developed a more generic SpeechNet2 network replacing the log mel-scale spectrogram feature pre-processor with a fully learnable convolution-based front-end. This front-end is designed to accept audio waveforms as inputs, which are apriori mu-law quantized on 8 bits to satisfy the input data type requirements of the ERGO processor.

## IV. SMART TOY PROOF-OF-CONCEPT

To demonstrate practical use cases for the DAVID platform we have developed two different example use cases. The original use case was a soft toy such as a Teddy bear that a child can talk with and engage in different play activities. A second example of a mobile robot was developed to take advantage of toy mobility to enhance some of the play activities, demonstrate additional functionality and explore if active mobility can help improve user engagement with the smart-toy. Here we present some technical details on how the underlying Edge-AI platform was integrated into each PoC.

### A. The DAVID Smart Teddy-Bear

The initial proof-of-concept (PoC) embodiment for the DAVID smart toy is a Teddy-Bear, or more correctly a Panda. This was chosen as many children have a special cuddly toy that they develop a close attachment to. Having a toy with a full speech interface that children can talk to provides an interesting toy variant to test and evaluate.

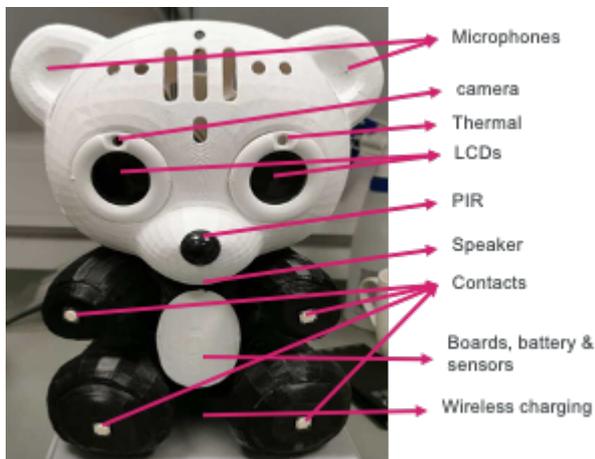

*Figure 8:* The DAVID Teddy Bear Design

From a technical perspective it is also less challenging to design and implement a static toy platform. For this proof-of-concept the focus is on providing a more sophisticated speech interface with interactive activities such as games and storytelling. The bear casing is designed to be 3D printable. A special 3D printing material is added later to provide a synthetic soft fur over the hard casing of the final toy. The toy platform hosts various sensing and UI elements as shown in *Figure 8*. It can be seated on a wireless charging cradle, but in normal operation it can run for more than 1 week on the internal, rechargeable battery pack.

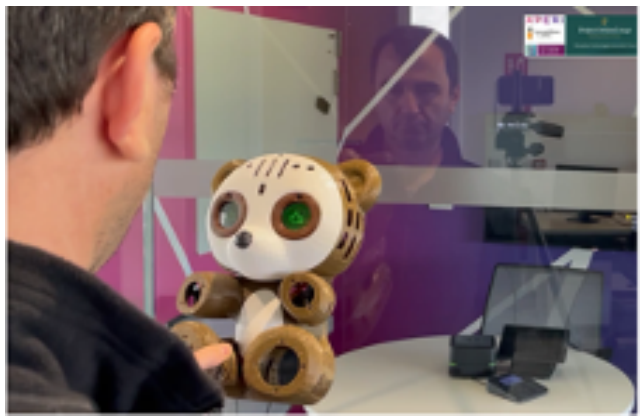

*Figure 9:* Testing Smart-Toy interactivity in the lab - the toy's eyes follow the user and can provide indications of emotional states.

### B. The DAVID Mobile Rover

A second design PoC embodiment is provided by a mobile rover. This is a pet-like platform that can move around and execute a variety of different movement patterns, including 180 and 360 degree rotations. It also adds a head-like data screen that can nod and antennae that can perform various movements. These can be used to express happiness or indicate the toy is upset or confused.

Several variants of this mobile design have been tested and improved. This platform can employ a less sophisticated speech interface for many demonstration tasks and games. Simple wake-words and simple speech commands can enable interesting play experiences. As this variant of the toy mimics a robo-pet the gameplay experiences can be simpler. Thus, chasing a ball, or dancing while playing some music can provide a very entertaining experience. Emotional cues can be simpler, again mimicking a pet puppy. Many of the demo activities for the mobile DAVID embodiment were noted by the project team as being more fun to work on than the sophisticated interactions expected in activities for the cuddly toy.

## V. CONCLUSIONS

The DAVID smart-toy platform provides an interesting overview of how future Edge-AI platforms are likely to evolve. The migration of much of the intelligence onto the sensing nodes can allow designers and developers to focus on the gaming or activity logic without worrying about how to integrate computer vision algorithms or speech recognition elements. Speech is analysed and generated from the underlying text representations and computer vision algorithms can provide authentication and analysis data in simplified forms.

Perhaps more important here are the data-privacy benefits. As both speech and face data are regarded as biometrics and thus classified as personally identifiable data (PID) developers would be exposed to GDPR compliance issues. By embedding the processing of this sensor data onto the ERGO node boards this removes the need for GDPR compliance for smart-toy designers.

REFERENCES

[1] M. A. Farooq, W. Shariff, and P. Corcoran, "Evaluation of Thermal Imaging on Embedded GPU Platforms for Application in Vehicular Assistance Systems," *IEEE Trans. Intell. Veh.*, pp. 1–1, 2022, doi: 10.1109/TIV.2022.3158094.

[2] D. Bigioi and P. Corcoran, "Challenges for edge-ai implementations of text-to-speech synthesis," in *2021 IEEE International Conference on Consumer Electronics (ICCE)*, IEEE, 2021, pp. 1–6.

[3] W. Yao, V. Varkarakis, G. Costache, J. Lemley, and P. Corcoran, "Towards Robust Facial Authentication for Low-Power Edge-AI Consumer Devices," *IEEE Access*, pp. 1–1, 2022, doi: 10.1109/ACCESS.2022.3224437.

[4] B. Sudharsan, "On-Device Learning, Optimization, Efficient Deployment and Execution of Machine Learning Algorithms on Resource-Constrained IoT Hardware," 2022.

[5] B. Sudharsan, S. Malik, P. Corcoran, P. Patel, J. G. Breslin, and M. I. Ali, "OWSNet: Towards Real-time Offensive Words Spotting Network for Consumer IoT Devices," in *2021 IEEE 7th World Forum on Internet of Things (WF-IoT)*, Jun. 2021, pp. 83–88. doi: 10.1109/WF-IoT51360.2021.9595421.

[6] B. Sudharsan, P. Patel, A. Wahid, M. Yahya, J. G. Breslin, and M. I. Ali, "Porting and execution of anomalies detection models on embedded systems in iot: Demo abstract," in *Proceedings of the international conference on internet-of-things design and implementation*, 2021, pp. 265–266.

[7] P. Pandey *et al.*, "Challenges and opportunities in near-threshold dnn accelerators around timing errors," *J. Low Power Electron. Appl.*, vol. 10, no. 4, p. 33, 2020.

[8] A. Reuther, P. Michaleas, M. Jones, V. Gadepally, S. Samsi, and J. Kepner, "AI and ML accelerator survey and trends," in *2022 IEEE High Performance Extreme Computing Conference (HPEC)*, IEEE, 2022, pp. 1–10.

[9] A. Reuther, P. Michaleas, M. Jones, V. Gadepally, S. Samsi, and J. Kepner, "Survey and benchmarking of machine learning accelerators," in *2019 IEEE high performance extreme computing conference (HPEC)*, IEEE, 2019, pp. 1–9.

[10] P. M. Corcoran, "A privacy framework for the Internet of Things," in *2016 IEEE 3rd World Forum on Internet of Things (WF-IoT)*, Dec. 2016, pp. 13–18. doi: 10.1109/WF-IoT.2016.7845505.

[11] P. Corcoran, "Privacy challenges for smart-cities: The challenge of IoT camera uberveilance," IEEE World Forum on Internet of Things, 2019.

[12] J. Hinds, E. J. Williams, and A. N. Joinson, "'It wouldn't happen to me': Privacy concerns and perspectives following the Cambridge Analytica scandal," *Int. J. Hum.-Comput. Stud.*, vol. 143, p. 102498, 2020.

[13] A. Karale, "The challenges of IoT addressing security, ethics, privacy, and laws," *Internet Things*, vol. 15, p. 100420, 2021.

[14] E. Taylor and K. Michael, "Smart toys that are the stuff of nightmares," *IEEE Technol. Soc. Mag.*, vol. 35, no. 1, pp. 8–10, 2016.

[15] J. Valente and A. A. Cardenas, "Security & privacy in smart toys," in *Proceedings of the 2017 Workshop on Internet of Things Security and Privacy*, 2017, pp. 19–24.

[16] H. Li, L. Yu, and W. He, "The impact of GDPR on global technology development," *Journal of Global Information Technology Management*, vol. 22, no. 1. Taylor & Francis, pp. 1–6, 2019.

[17] "Perceive – Transform Sensing into Perceiving." https://perceive.io (accessed Jul. 14, 2023).

[18] P. Corcoran, P. Bigioi, E. Steinberg, and A. Pososin, "Automated in-camera detection of flash-eye defects," *IEEE Trans. Consum. Electron.*, vol. 51, no. 1, pp. 11–17, 2005.

[19] P. M Corcoran, P. Bigioi, and F. Nanu, "Advances in the detection & repair of flash-eye defects in digital images-a review of recent patents," *Recent Pat. Electr. Electron. Eng. Former. Recent Pat. Electr. Eng.*, vol. 5, no. 1, pp. 30–54, 2012.

[20] N. Kaur and P. Singh, "Conventional and contemporary approaches used in text to speech synthesis: a review," *Artif. Intell. Rev.*, vol. 56, no. 7, pp. 5837–5880, Jul. 2023, doi: 10.1007/s10462-022-10315-0.

[21] R. Badlani, A. Łańcucki, K. J. Shih, R. Valle, W. Ping, and B. Catanzaro, "One TTS alignment to rule them all," in *ICASSP 2022-2022 IEEE International Conference on Acoustics, Speech and Signal Processing (ICASSP)*, IEEE, 2022, pp. 6092–6096.

[22] J. Li *et al.*, "Jasper: An end-to-end convolutional neural acoustic model," *ArXiv Prepr. ArXiv190403288*, 2019.

[23] S. Kriman *et al.*, "Quartznet: Deep automatic speech recognition with 1d time-channel separable convolutions," in *ICASSP 2020-2020 IEEE International Conference on Acoustics, Speech and Signal Processing (ICASSP)*, IEEE, 2020, pp. 6124–6128.

[24] C. Szegedy, S. Ioffe, V. Vanhoucke, and A. Alemi, "Inception-v4, inception-resnet and the impact of residual connections on learning," in *Proceedings of the AAAI conference on artificial intelligence*, 2017.

[25] A. Graves, S. Fernández, F. Gomez, and J. Schmidhuber, "Connectionist temporal classification: labelling unsegmented sequence data with recurrent neural networks," in *Proceedings of the 23rd international conference on Machine learning*, 2006, pp. 369–376.